# Consistency of Feature Attribution in Deep Learning Architectures for Multi-Omics

Daniel Claborne, Javier Flores, Samantha Erwin, Luke Durell, Rachel Richardson, Ruby Fore, Lisa Bramer.

*Pacific Northwest National Laboratory*

## Abstract

Machine and deep learning have grown in popularity and use in biological research over the last decade but still present challenges in interpretability of the fitted model. The development and use of metrics to determine features driving predictions and increase model interpretability continues to be an open area of research.  We investigate the use of Shapley Additive Explanations (SHAP) on a multi-view deep learning model applied to multi-omics data for the purposes of identifying biomolecules of interest.  Rankings of features via these attribution methods are compared across various architectures to evaluate consistency of the method.  We perform multiple computational experiments to assess the robustness of SHAP and investigate modeling approaches and diagnostics to increase and measure the reliability of the identification of important features.  Accuracy of a random-forest model fit on subsets of features selected as being most influential as well as clustering quality using only these features are used as a measure of effectiveness of the attribution method.  Our findings indicate that the rankings of features resulting from SHAP are sensitive to the choice of architecture as well as different random initializations of weights, suggesting caution when using attribution methods on multi-view deep learning models applied to multi-omics data.  We present an alternative, simple method to assess the robustness of identification of important biomolecules.

## Introduction

Improvements in instrumentation and decreased costs have led to an increased ability to generate a larger number of samples in omics-based biological studies. As a result, machine learning, and in particular, deep learning (DL) has emerged as an effective method for integrating multi-omics datasets for tasks such as cancer subtype prediction [1], [2], [3] and survival analysis of cancer patients [2], [6].  These DL methods for multi-omics data often rely on architectures that jointly ingest data from several omics sources (views) before combining their representations and performing a classification task [4], [5], [6], [7].  Although these methods have demonstrated the ability to perform well on tasks such as classification of treatment groups, based on multi-omics data, model interpretability (e.g. identifying biomarkers of a treatment group) is of great importance in biological applications. The black box nature of deep neural networks makes them difficult to use when trying to gain insight into the biological system. Shapley Additive Explanations (SHAP) have emerged as a popular method to increase DL model interpretability and for trying to explain which features drive predictions.

For example, Yap et al. [8] used SHAP to identify genes that were predictive of tissue types in genotype-tissue expression RNA-seq data.  The most important genes determined by SHAP were found to be biologically relevant for tissue differentiation processes and performed better on clustering tasks as compared to all genes or a random subset. In another example,  Benkirane et al. [2] used SHAP towards their multi-view variational autoencoder to identify genes in RNA-seq data that most affected predictions of breast cancer subtypes.  Some well-known biomarkers of breast cancer were among the most important genes identified by SHAP. As a final example, Tasaki et al. [9] applied SHAP scores to their deep learning model that was fitted to predict differential expression of gene transcripts from various complex molecular interactions such as transcription factor promoter interactions. The SHAP scores allowed the authors to compare the importance between different interaction types and perform an enrichment analysis that identified biological processes known to be associated with gene regulation.

The SHAP scores used in each of these examples inherits ideas from Shapley values, a concept from game theory in which the marginal contribution of each 'player' on a team contributes to the final payout. This concept has been translated to machine learning by considering each feature as a player and the model prediction as the payout.  While not directly applicable to deep learning, the method of Shrikumar et al. [10] has been shown to be able to approximate Shapley values within a deep learning framework [11].  In this work we use "SHAP" to specifically refer to the modified version of Shrikumar et al. found in the DeepExplainer class of the SHAP python package first presented in work by Lundberg & Lee [11].

This work focuses on 'multi-view' networks that are popular in multi-omics deep learning applications [2], [3], [4], [6], [12], [13], [14], [15] . Given the popularity and use of SHAP for elucidating the predictive drivers within multi-view deep learning models and as a tool for identifying drivers of biological outcomes, it is important that SHAP's capabilities and limitations be fully assessed. Therefore, we conducted and present experiments evaluating the robustness of SHAP (i.e. variable importance) to varying modeling choices.  Ideally, the importance of a fixed set of biomolecules should remain relatively stable across different architectures, excluding degenerate architectures that effectively ignore some of the inputs or are otherwise unsupported by the data.  We investigate the reliability of SHAP within architectures that are common in practice and the literature.  Additionally, deep learning commonly includes various random components during training such as initialization of the weights, randomized batching of training examples, and random sampling of intermediate features. We further evaluate the size of the discrepancy in SHAP importance due to such sources of randomness in the DL modeling process. Understanding the effect of these random components is important to establishing a more reliable protocol for identifying important features that considers this randomness to help prevent false categorization of features as important or unimportant.

# Methods

## Model Architecture

Multi-view networks' architectures commonly involve feeding measurements from several layers of the omics hierarchy into separate models that consist of one or more standard fully connected layers.  Hereafter, we refer to these per-omic-type models as the marginal models. These marginal models output both a vector representation (latent vector) of their inputs and predicted probabilities of the response categories for each input sample.  After each input is processed through its respective marginal model, the latent vector outputs are combined in one of two ways:  i) averaging, if vector dimensions are equivalent across all views, or ii) concatenating if the vectors are of differing dimension.  The combined outputs are then ingested by a set of final prediction layer(s). The template of this architecture is represented in Figure 1.

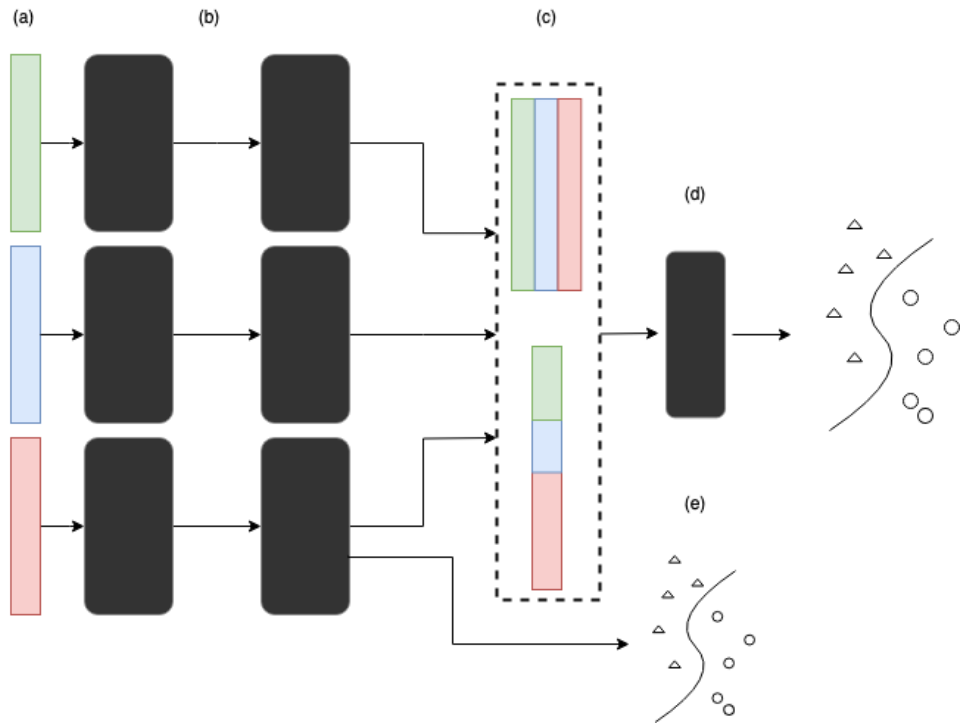

*Figure 1:  High level overview of our model architecture.  (a) Separate 'views' or `omics types to be input to separate marginal networks.  (b) Two hidden linear layers with ReLU activations per marginal network, each processing a single view.  (c) Two possible combination schemes:  The equal-dimension outputs of each marginal network are averaged feature wise (top scheme), or possibly non-equal size outputs are concatenated (bottom scheme).  (d) Final hidden linear layer processes the combined representation from (c) before classification.  (e) Optional per-marginal network classification task, shown here for a single marginal network.*

Our architecture is a simplification of the architecture proposed in Lee et. al. and more directly conforms to the base template in Figure 1 that generally characterizes other similar multi-view networks. The choice of this simplified architecture is motivated by the desire to easily perform ablation experiments within the general multi-view framework. In this way we also avoid having to deal with discrepancies resulting from unique modeling choices of various multi-view implementations. For example, our architecture avoids some instability associated with the variational loss when estimating model parameters in Lee et al. The loss used by Lee et al. contains a Kullback-Leibler (KL) divergence term primarily used in the machine learning literature to train generative models, and other authors have noted issues with using KL divergence as a regularization term [2].

We use feed-forward neural networks for each of our marginal models, rather than the stochastic encoders implemented in Lee et al. We use between two and three marginal models, depending on the number of views in an experiment, each with 2 hidden layers (consisting of a linear projection followed by a nonlinear activation function) of varying size and a prediction head for per-marginal model predictions. Like Lee et al., the output of our marginal models are mappings to some lower-dimensional space, but they are not defined by some probabilistic distribution. We combine these marginal outputs via averaging or concatenation before feeding them through another hidden layer and a final prediction head. Using the averaging approach, our model can also handle the presence of completely missing views, as is the case in data from The Cancer Genome Atlas Pan-Cancer (TCGA) project [16]. If a view is missing, we simply compute the average of the outputs from the marginal models with non-missing views.

Our loss is the focal loss from Lin et al. [17], an adjustment to standard categorical cross-entropy for classification tasks that further penalizes incorrect predictions and deprioritizes predictions that are already confidently correct. The datasets used in our experiments are small enough that we perform full gradient descent using all samples from each dataset at each iteration, and we use the Adam optimizer for it's good out-of-the-box performance [18]. In our experiments we use a validation set to observe a plateau in performance for the purposes of selecting the final number of training iterations. Outside of training iterations and the architectural choices previously mentioned, we do not tune hyperparameters such as learning rate, focal loss constants, or dropout. To ensure our simplified architecture is reasonable we compared its performance to that of Lee et al. on a 1-year mortality prediction task on TCGA data and found them to perform similarly. All experiments in this manuscript are therefore based on our simplified model, however one of our main experiments is repeated based on Lee et al.'s model to validate the generalizability of observed patterns. Details of the model comparisons and the experimental results based on Lee et al.'s model are available in the supplement.

## Data

We use two multi-omic datasets (i.e. ICL102 and ICL104) throughout our experiments, both derived from larger compendium studies performed by the 'Omics of Lethal Human Viruses (OMICS-LHV) Systems Biology Center and funded by the National Institutes of Allergy and Infectious Diseases (NIAID) (grant # U19AI106772) from 2013 to June 2018.

The ICL102 data were originally collected as part of a study to evaluate human host cellular response to Influenza A virus (subtype H7N9): wild-type strain Influenza A/Anhui/1/2013 (AH1-WT), mutant viruses NS1-L103F/I106M (AH1-F/M), and partially ferret-adapted (AH1-691) infection. Sample data was obtained from human lung adenocarcinoma cells (Calu-3) and consist of proteomics (# proteins = 3722), lipidomics (positive and negative ionization modes; # lipids = 496), and metabolomics data (# metabolites = 80; 50 of which were annotated with known labels) measured on five control samples, five wild-type strain Influenza A/Anhui/1/2013 (AH1-WT) infected samples, five mutant virus NS1-L103F/I106M (AH1-F/M) infected samples, and five partially ferret-adapted (AH1-691) infected samples across six timepoints (0-, 3-, 7-, 12-, 18-, and 24-hours post-infection), for a total of 120 sample measurements.

The ICL104 data were originally collected as part of a study to evaluate host response to pandemic Influenza A virus (subtype H1N1), natural isolate Influenza A/California/04/2009 virus. Sample data was obtained from human lung adenocarcinoma cells (Calu-3) and consist of proteomics (# proteins = 4896), lipidomics (positive and negative ionization modes; # lipids = 491), and metabolomics data (# metabolites = 138; 72 of which were annotated with known labels) measured on five control samples and five A/influenza/California/04/2009 (H1N1-CA04) infected samples across six timepoints (0-, 7-, 12-, 24-, 36-, and 48-hours post-infection), for a total of 60 sample measurements. Further details on these studies and their experimental protocols may be found in Eisfeld et al [19].

## Ablation Experiments

We perform a series of ablation experiments to explore the behavior of SHAP outputs across several sources of variation: view input size, layer size, and randomness during training (See supplementary items 1-5). Specifically, our interest is in determining i) whether the number of biomolecules in each input view artificially affects the overall importance of features from that view; ii) the ability of layer size choices to consistently control variation in SHAP scores; iii) the variation in SHAP scores due to different final trained models resulting from random weight initialization and dropout; and iv) the performance of selected features in classification and clustering tasks across different architectures. Details of the performed experiments are provided in the following subsections.

## Effect of Feature Compression on SHAP Values

Multi-omics datasets commonly have views with very large numbers of features alongside views with fewer features, as seen in the ICL104/ICL102 proteomics versus the lipidomics and metabolomics datasets. Deep learning architectures are often designed to create a 'compressed' representation of their inputs, reducing an input of thousands of values into a single vector/embedding of perhaps, 32, 64, or 128 values. A question follows about how much the level of compression (e.g. reducing 1000 input features to a 100-dimensional versus 10-dimensional representation) affects SHAP scores, and more specifically to what extent different levels of compression across multiple views affects rankings based on SHAP scores. To answer this question, we experiment with adding extra noise features to the input as well as adjusting the relative size of the final embedding layer of the marginal models.

First, we use the metabolomics and proteomics data from the ICL104 dataset in our multi-view setup with 2 marginal models, sequentially adding sets of randomly generated noise features to the metabolomics data to investigate the effect of input view size on overall importance. The number of features in the proteomics data is held fixed throughout all experiments so that we can determine whether the importance values of the real features in the metabolomics data augmented with noise are reduced relative to those in the other view. Added noise features are generated based on a normal distribution with mean and variance sampled from the sample means and variances of the original metabolomics features. We add $n = \{0, 100, 500, 1000, 2500, 4000\}$ noise features to the original metabolomics feature set and train a separate network with the augmented metabolomics data and original proteomics dataset for each case. We then measure both the change in magnitude of SHAP scores and change in rankings of the original features, with and without added noise. For magnitude, we average the absolute SHAP values. For comparing if two sets of rankings are similar, we use weighted Kendall's Tau [20] which compares the rank-ordering of a returned set of two scores but assigns more weight to matches at the top ranks (most important features). For each noise level, 10 training runs are performed to get a sense of the distribution of magnitude and Kendall's Tau values.

We also test varying the amount of compression via layer size, since this is often more directly controllable than the input size and perhaps can be intelligently adjusted to achieve the right balance of compression across marginal models in a way that input size cannot. To test the ability of layer sizes to control degradation in SHAP values, we begin with a 'base' model size, which we instantiate as a two-layer network with sizes (64,64) for the metabolomics marginal model and either (128, 128) or (512, 512) for the hidden layers of the proteomics marginal model, chosen to represent common sizes seen in practice. We then add noise features to the metabolomics data and train two models from scratch: One with the base layer sizes, and another where we adjust the size of each marginal network (all layers, including the combination layer) to account for the change in input size (i.e. the metabolomics network increases in size relative to the proteomics network). This process is repeated across the same increasing levels of noise.

Ideally, we would like to see a predictable effect of marginal model output layer size on importance, allowing recommendations for how to size these layers to make comparisons across views 'fair'. Additionally, we want to see any such adjustment does not decrease classification performance, and we describe methods to test this in the section on performance of feature subsets. We provide the exact layer sizes, noise level, and other parameters for these experiments in supplementary items 1-5.

## Variation of Importance Values Across Training Runs

We also investigate the variation in ranks of specific features resulting from randomness across training runs; in our experiments, this variation is due to the random initialization of weights in the network and dropout layers during training. Other common sources of random variation include random sampling of training batches, though our data is small enough that we can perform gradient descent on the full data. Our motivation for this experiment is to see how non-user-controlled sources of variation affect SHAP values for specific biomolecules, in contrast to variation due to explicit selections such as layer size. We use the same output as the feature compression experiments and see whether particular features can be characterized by their propensity to stay in high or low ranks.

## Performance of Features Subsets

Finally, we test how sensitive model performance is to changes in layer size and combination scheme (mean vs concatenation) in terms of classification and clustering performance metrics (not consistency of features). For these experiments, we use all 3 views from both the ICL104 and ICL102 datasets and determine layer sizes by choosing varying dimensions at the combination layer and then assigning intermediate layer sizes between the input dimension for each biomolecule input and the combination layer size. In the mean combination approach, the combination layer size must be the same for all 3 marginal models. For the concatenation combination scheme, we choose several combination layer sizes for the view with the smallest input dimension (metabolites). The combination layer size for the other views is then chosen to be larger by the same proportion as its input dimension is to the smallest input.

To compare each architecture, we train each model until we observe a plateau in validation performance and then retrain the model for 20% more iterations on the combined training + validation set. For each model, SHAP values are computed on the final training data. The average of the absolute value of each score across all samples is used as the final score for each biomolecule. We take the top $p$ percent of biomolecules as determined by the computed SHAP score and use them to fit a random forest classifier; we tested $p = \{75, 50, 25, 10\}$. Performance is measured relative to a random forest model fit on all the features. We choose random forest for its reputation for good 'out-of-the-box' performance [21], [22]. Additionally, we compare across architectures by using the same process of selecting the top $p$ percent biomolecules as determined by importance metrics from each model on classification and clustering tasks. For classification, we report area under the receiver operating curve (AUC) and for clustering we use the V-measure [23] for cluster quality.

Exact hidden layer sizes for each of the marginal models and combination layer, and performance of our model and random forest models on a holdout set are given in the supplementary material.

# Results

## Feature Compression

For our noise-augmentation experiments we find that the noise features interfere with both the magnitude and the rankings of the original features. Weighted Kendall's Tau decreases consistently as more noise features are added, as can be seen in Figure 2. A value of 1 indicates perfect matching of ranks between two setups, while a value of -1 indicates completely opposite ranks, and 0 indicating no correlation. We see generally high values of weighted Kendall's Tau, however the difference in drop-off is clear. This suggests that rankings of features that together effectively discriminate between classes can be disrupted by inflation of the input size.

Figure 2 shows Kendall's Tau between rankings of SHAP values resulting from models fit on the original data (zero noise features) versus the augmented data. In the blue boxplots of a model beginning with hidden layer sizes of 128, we see that as noise increases, and if we keep the layer sizes fixed, the Shapley values start to diverge from the original. If we dynamically change the layer size as described in the methods section, we can see that the Shapley values are much more stable (orange boxplots). This is promising, unfortunately, the test performance of each model is not stable in this example; test performance of the models with dynamic layer sizes decreases as noise increases, while those of the fixed model remain near-perfect. This may be caused by the fact that the scheme for modifying layers can result in very small layer sizes for the marginal network that ingests the un-augmented proteomics data, small enough that they do not have enough capacity to learn all the relationships in that view. This phenomenon of degrading correlation can be seen in the architecture of Lee et al. as well (See the figure of supplementary item 6).

Further, the observed behavior of degrading SHAP values is not consistent across all base hidden layer sizes. The red and green boxplots in Figure 2 show what happens when the network of the fixed dataset begins at a larger size (two hidden layers of size 512). The pattern is reversed in this case – the fixed model is now the one with more stable SHAP values. Additionally, for the fixed architecture, though we still see some degradation, the drop-off in Kendall's Tau is less pronounced than in the smaller fixed architecture. The layer size for the network ingesting the augmented data is the same in both cases, so the instability in correlation values is driven by the larger layer sizes of the second network.

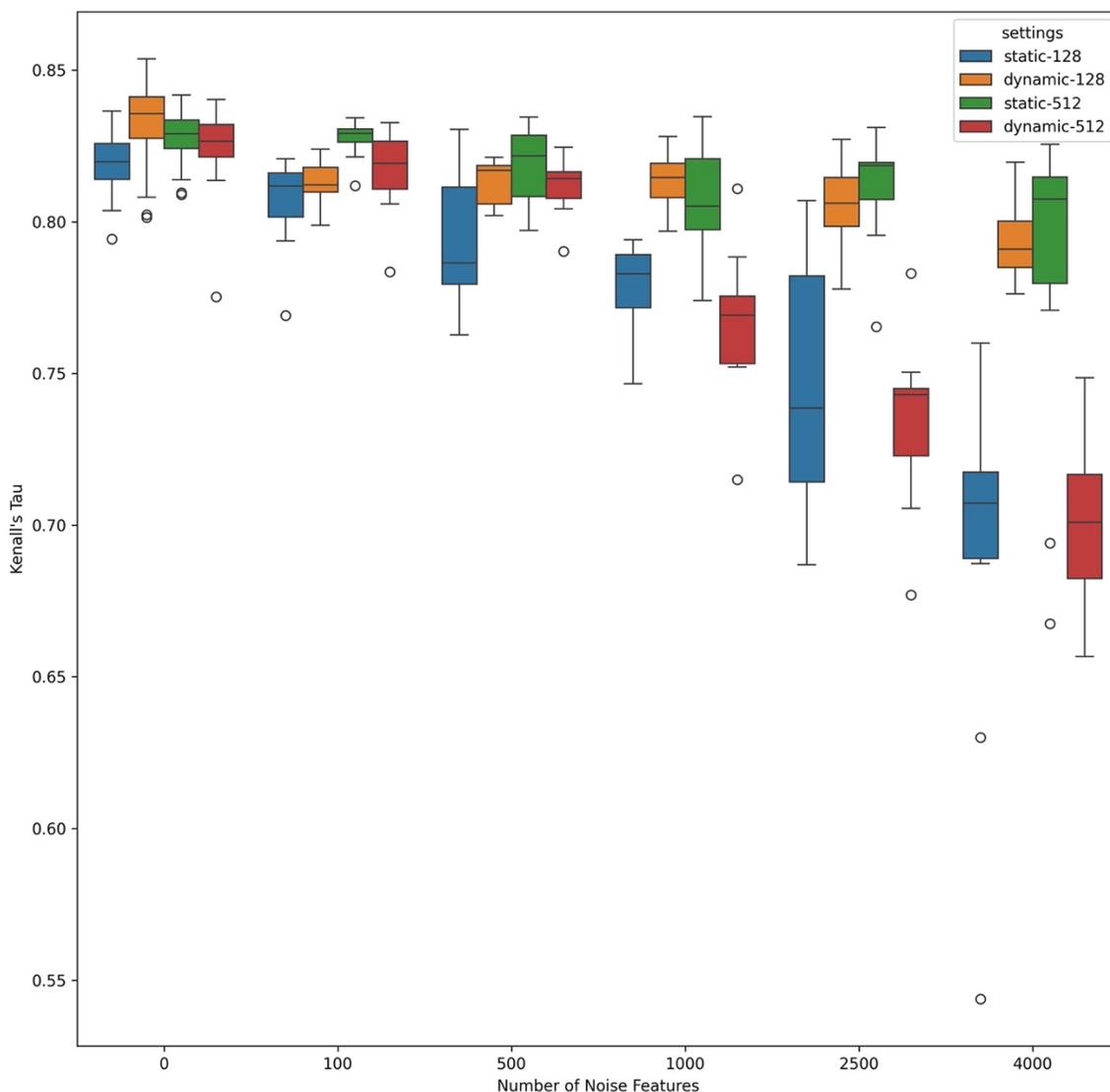

*Figure 2: Kendall's Tau between biomolecule rankings for different levels of noise using SHAP values to determine ranks. Each box and whiskers represent 10 training runs at a particular noise level and layer size scheme: 'static' refers to the scheme where layers do not change hidden layer size as noise level increases, 'dynamic' refers to the scheme where we increase the layer size of the marginal model whose input is being augmented and decrease the size of the other model. '128' and '512' refer to the initial sizes of the hidden layers of the second (un-augmented) marginal model (initial sizes of the hidden layers of the first marginal model are always 64).*

## Variation in Features

We find significant variation in the ranks of real features across training runs. Figure 3 shows distributions of ranks across training runs for the top (i.e. ranks 1,2,3..) and bottom (i.e. ranks 351, 350, ... out of 351 possible ranks) 8 ranked biomolecules as determined by average rank across training runs. Even for the top ranks, we see that they can range from the top 1% of ranks down towards middle ranks. They are however, fairly concentrated among the 'upper' ranks. By contrast the low ranked biomolecules are concentrated around lower ranks. Of note are the biomolecules that only occasionally achieve high ranks but mostly occupy very low ranks. A given training run might suggest one of these features is important, even though across many training runs it will rarely be identified as a top driver of predictive performance. It seems reasonable to categorize these features as less important than those that maintain high ranks more consistently.

We inspected the change in ranks when considering only the SHAP scores of the real features in the metabolomics dataset, as well as when considering all SHAP scores from both the metabolites and second omics (lipids or proteins). When considering only the metabolites, the pattern of variation in SHAP scores does not seem to be affected by added noise (See supplementary item 7). Given that we showed that rank correlation values between original and augmented data degrade as the number of noise features is increased, we suspect that while the noise may affect the rank correlation by 'shuffling' ranks, biomolecules tend to retain their general distribution of ranks among all features.

When considering SHAP scores from both views together, as is the case in Figure 3, we see the ranks of top ranked biomolecules trend towards lower ranks as noise is added. This is the same phenomenon as observed in the blue boxplots of Figure 2, that even for top ranked biomolecules, spurious features in one view can cause important biomolecules in that view to score lower relative to biomolecules in another view.

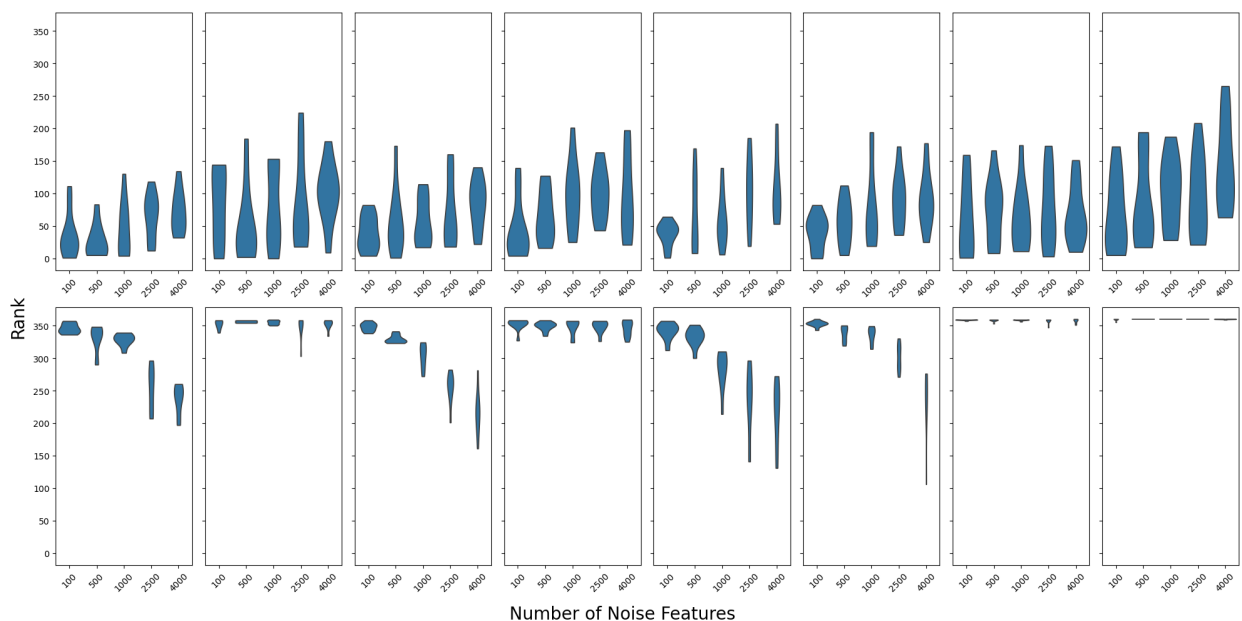

*Figure 3: Effect of noise on rank for the highest (top row) and lowest (bottom row) 8 ranked metabolites using the dynamic layer sizing scheme. Ranks are computed among features for both views to see the effect on their global ranking.*

## Binary and Multi-class Viral Dataset Classification Performance

On the binary classification viral dataset, we see comparable performance between the two combination schemes. The binary dataset has a much easier classification task, with perfect AUC on a holdout set across every run. Using SHAP values to subset to the top 10% of features and refit a random forest resulted in high AUC scores, but with a quarter of the runs dropping below perfect for the mean combination approach and half below perfect for concatenation.

The multiclass viral dataset is unsurprisingly more difficult, with AUC scores below perfect on all runs. Mean fusion averaged 0.71 AUC while concatenation fusion averaged 0.76 across 16 runs each. For the subsetting experiments, performance appears to be maintained by both schemes. Subsetting to 10% based on top features based on Shapley values resulted in 0.75 and 0.71 AUC for the mean and concatenation fusion approaches. This result is somewhat more surprising, suggesting that most discriminative power is contained within the top features for this dataset.

In both the ICL102 and ICL104 datasets, clustering performance as measured by V-measure was quite poor. Across all subsetting levels we do not see V-measure scores above 0.1 (V-measure ranges from 0 to 1). We do however see a decrease in V-measure score as we reduce the proportion of top features selected.

All performance metrics for both datasets and all layer sizes/combination schemes are given in the supplementary material, specifically items 3 and 4.

## Discussion

Our experiments generally show that SHAP values, when applied in a simple deep learning setting can be highly variable. When many noise features are present, the ranking of other features that strongly predict the output in the absence of the noise features can be distorted. Skewing the layer size in accordance with increased noise features showed inconsistent ability to control this effect. The scheme with smaller layer sizes showed promising results in stabilizing variable attribution, but poor stability in actual predictive performance. A similar scheme with the second network being of larger size showed stable predictive performance but limited effect in controlling the consistency of variable attribution.

We speculate that the degradation induced by noise is due to deep network's strong ability to find any combination of features which can reduce training error, and as the number of noise features increases, the possibility that it can erroneously exploit one of those features, possibly in combination with a feature that is driving the variation in the output, increases. As for the 'flipping' of results when the size of the second network changed, it is possible that a smaller network constrains the ability of the network to erroneously exploit noise features, making the results more consistent across training runs.

Variation in the ranks of features across training runs due to initialization of weights is higher than we expected, with even some top ranked or lower ranked biomolecules occasionally flipping to the opposite side of the ranking hierarchy for some training runs. We still believe that visual inspection of the variation in ranks gives a better sense of whether a biomolecule is truly 'important'.

Though the consistency of the returned features may be poor, our experiments comparing mean and concatenation combination show that the discriminative quality of the returned features remains consistent when passed downstream to a simpler model such as random forest or a clustering task.  This is perhaps not surprising for random forest, as it has a similar ability to identify complex, though possibly erroneous, interactions.  One might expect a clustering task with agglomerative clustering to degrade, since it is based on simple distance metrics between observations, but we saw no significant drop as measured by V-measure.  It is possible that our base datasets already contained highly nonlinear relationships and were difficult for the clustering task regardless of which set of features was used to obtain the result.

# Conclusion and Future Work

We applied the popular variable attribution method Shapley Additive Explanations (SHAP) to multi-view deep networks for multi-omics data to probe the resulting scores and associated rankings of biomolecules.  Our results demonstrate that this popular method shows great variability in which biomolecules are indicated as most important for driving discriminative performance in a classification task.  The variation comes both from architectural choices such as layer size and combination scheme, as well as the random initialization of weights when beginning training of deep neural networks.  Biomolecules that are flagged as most important due to a high SHAP score in one run, can sometimes be ranked as below median importance in another run with different initialization.  Additionally, the size and density of important features can cause shuffling among ranks, as suggested by experiments in which noise features were introduced into one view.  Our experiments overall suggest caution when using the output of variable attribution methods such as SHAP in deep learning frameworks, and if computing resources allow, to obtain scores from multiple training runs and inspect their variation when identifying molecular candidates for determining drivers of biological outcomes.

Future research towards the effectiveness of variable importance metrics could inspect the biological plausibility of features selected through visual inspection of the variation in scores across runs.  Further investigation into reliable methods for remedying the observed variation in SHAP values is warranted.

# Data Availability

The raw data from which our data was derived and transformed is available from Pacific Northwest National Laboratory's DataHub page.  The ICL102 dataset is hosted here: https://data.pnnl.gov/group/nodes/dataset/13112, and the ICL104 dataset is hosted here: https://data.pnnl.gov/group/nodes/dataset/13114.

The wrangled/processed data analyzed during the current study are available from the corresponding author on reasonable request.

# Funding

This work was funded by the Defense Threat Reduction Agency.

# Acknowledgements

The Pacific Northwest National Laboratory is a multi-program laboratory operated by Battelle for the U.S. Department of Energy under contract DEAC05-76RL01830.